\documentclass{article}

\usepackage{microtype}
\usepackage{graphicx}
\usepackage{subfigure}
\usepackage{booktabs} 
\usepackage{placeins}
\usepackage{multirow}
\usepackage{chemformula}
\usepackage[normalem]{ulem}

\usepackage{hyperref}

\usepackage[accepted]{icml2025}

\usepackage{amsmath}
\usepackage{amssymb}
\usepackage{mathtools}
\usepackage{amsthm}

\usepackage[capitalize,noabbrev]{cleveref}

\theoremstyle{plain}

\theoremstyle{definition}

\theoremstyle{remark}

\usepackage[textsize=tiny]{todonotes}

\icmltitlerunning{Differentiable Chemistry in PINNs for Solving Parameterized and Stiff Reaction Systems}

\begin{document}

\twocolumn[
\icmltitle{Differentiable Chemistry in Physics-Informed Neural Networks for Solving Parameterized and Stiff Reaction Systems}


\icmlsetsymbol{equal}{*}

\begin{icmlauthorlist}
\icmlauthor{Miloš Babić}{yyy,zzz}
\icmlauthor{Franz M. Rohrhofer}{comp}
\icmlauthor{Stefan Posch}{yyy,zzz}
\end{icmlauthorlist}

\icmlaffiliation{yyy}{CD Laboratory for Physics-driven Machine Learning in Industrial Applications, Graz, Austria}
\icmlaffiliation{comp}{Know Center Research GmbH, Graz, Austria}
\icmlaffiliation{zzz}{The Institute of Thermodynamics and Sustainable Propulsion Systems, Graz University of Technology, Graz, Austria}

\icmlcorrespondingauthor{Miloš Babić}{milos.babic@tugraz.at}

\icmlkeywords{Machine Learning, ICML}

\vskip 0.3in
]



\printAffiliationsAndNotice{}  

\begin{abstract}
From neural ODEs to continuous-time machine learning, differentiable solvers allow physics, optimization, and simulation to become trainable components within deep learning systems.
This has opened the path to a new generation of deep learning frameworks for scientific computing, with many promising applications still emerging.
In this paper, we integrate a differentiable chemistry solver into a modified physics-informed neural network to solve parameterized reaction systems that are inherently stiff.
The proposed framework introduces several key components required to overcome limitations of standard physics-informed neural networks.
These include a differentiable chemistry solver, a network architecture for parameterized solutions, and residual weighting tailored to stiff reactions. 
We evaluate the framework on a set of differential equations related to hydrogen combustion, which include initial/boundary value problems, inverse parameter identification, and a parameterized partial differential equation.
Our results highlight the ability of the proposed approach to extend physics-informed neural networks to stiff chemical systems that were previously inaccessible.
\end{abstract}

\section{Introduction}

Physics-informed machine learning (PIML) has emerged as a powerful paradigm in scientific computing, combining data-driven learning with prior knowledge derived from physical laws \cite{karniadakis2021physics}. By embedding known physical constraints directly into modern machine learning models, PIML aims to improve generalization, data efficiency, and physical consistency in the modeling of complex systems. Over the past decade, a growing body of work has demonstrated that purely data-driven approaches often struggle to achieve the accuracy and robustness required for challenging physical problems, motivating the integration of physics into the learning process.

Early developments such as Hamiltonian Neural Networks~\cite{greydanus2019hamiltonian}, Lagrangian Neural Networks~\cite{cranmer2020lagrangian}, and Neural Ordinary Differential Equations~\cite{chen2018neural} have highlighted the importance of respecting underlying physical structure during learning. Among these approaches, physics-informed neural networks (PINNs) have become a central workhorse of PIML. PINNs encode governing equations, constitutive relations, and constraints directly into the loss function through physics-based residual terms, enabling the solution of forward and inverse problems governed by differential equations~\cite{RAISSI2019686,cuomo2022scientific}.

Since their introduction, PINNs have attracted significant attention and inspired numerous extensions that build upon the same principle of physics-based loss formulations. These developments have pushed the boundaries of scientific machine learning toward problems that were traditionally considered intractable using classical numerical methods alone. Notable examples include deep operator learning approaches such as DeepONets \cite{lu2021learning}, Fourier Neural Operators \cite{lifourier}, and physics-informed neural operators \cite{li2024physics}, which aim to learn entire solution operators rather than individual realizations of physical systems. These operator-learning frameworks have been successfully applied to a wide range of systems, including parametric fluid dynamics problems \cite{lifourier,kovachki2023neural}, stochastic and uncertainty-aware PDEs \cite{lu2021learning,zhu2023learning}, and multiscale modeling in materials science and solid mechanics \cite{koric2024deep,rezaei2025finite}.

For PIML reaction–diffusion systems constitute a particularly challenging class of physical systems, describing the interplay between chemical reactions and transport processes such as diffusion. These systems arise in numerous scientific and engineering applications, including combustion, catalysis, biological pattern formation, and materials science. Depending on the underlying reaction mechanisms and involved species, reaction–diffusion systems can become highly multi-scale and exhibit stiff dynamics, characterized by rapid temporal variations over very short time scales. Accurately resolving such behavior often requires specialized numerical solvers, or even operator splitting approaches that couple solvers tailored to fast chemical kinetics with those designed for slower transport processes~\cite{sportisse2000analysis}.

For systems exhibiting stiff dynamics, standard PINN formulations face well-documented limitations, including optimization difficulties, poor convergence, and imbalanced training across temporal or spatial domains~\cite{krishnapriyan2021characterizing,monaco2023training}. To address these challenges, several modifications to the original PINN framework have been proposed, each targeting specific failure modes such as spectral bias~\cite{wong2022learning,chai2024overcoming}, stiffness~\cite{daw2022rethinking,ji2021stiff}, or training imbalance~\cite{wang2021understanding,maddu2022inverse}. Despite these advances, existing PINN studies on reaction–diffusion systems have largely focused on simplified reaction terms, such as Fisher’s equation~\cite{pmlr-v235-cho24b,rohrhofer2025approximating} or related low-dimensional problems~\cite{rao2023encoding,krishnapriyan2021characterizing}. Prior efforts applying PINNs to combustion are valuable but do not address the unsteady, forward reaction–diffusion setting~\cite{wu2025}. They typically target steady configurations~\cite{cao2024}, inverse parameter inference~\cite{LIU2024, LIU2025}, or neglecting the chemistry in the loss term formulation~\cite{BODE2023}.

Recent advances in differentiable programming and differentiable numerical solvers provide a key opportunity to overcome these limitations~\cite{holl2024, FRANZ2026, hu2019, freeman2021}. In particular, differentiable chemistry solvers enable backpropagation through complex, stiff reaction kinetics, allowing chemical source terms to be incorporated into physics-based loss functions in an end-to-end differentiable manner. This capability fundamentally expands the scope of PINNs, enabling the seamless integration of high-fidelity chemistry models into neural network training without sacrificing physical accuracy or consistency.

In this work, we present a practical, end-to-end differentiable PINN framework for stiff combustion systems and demonstrate its effectiveness on hydrogen flamelet equations, which we propose as a challenging benchmark for future PINN research. Our framework introduces several essential modifications to the standard PINN formulation that are required to achieve accurate and stable solutions in the presence of stiff chemical kinetics. These include a fully differentiable chemistry solver, residual-based loss weighting strategies, and the enforcement of hard physical constraints such as mass conservation, boundary and initial conditions, and inert species mass fractions.

We evaluate the proposed framework on a range of problem settings for hydrogen combustion, including both ordinary and partial differential equations. The considered tasks span forward problems—such as initial value problems, boundary value problems, and operator learning—as well as an inverse problem involving parameter identification. Furthermore, we compare our approach against the standard PINN framework and against a state-of-the-art enhancement technique for time-dependent problems which utilizes causality-based loss scaling~\cite{wang2022respecting}. Our results demonstrate that the proposed differentiable PINN framework significantly improves robustness and accuracy for stiff reaction–diffusion systems, highlighting the critical role of differentiable solvers in enabling next-generation physics-informed learning.

In summary, our main contributions are as follows:
\begin{itemize}
    \item We design a physics-informed neural network framework that integrates a differentiable chemistry solver, together with additional components required to address stiff reaction–diffusion systems.

    \item We evaluate the proposed framework on a diverse set of problem definitions, including ODEs and PDEs, covering forward problems (initial and boundary value problems) as well as an inverse problem for parameter identification.

    \item We extend the framework to parameterized partial differential equations, demonstrating its effectiveness for operator learning in chemically reacting systems.

    \item We conduct an ablation study to systematically assess the relevance and impact of each framework component.
\end{itemize}

\begin{figure*}[ht]
  \vskip 0.2in
  \begin{center}
    \centerline{\includegraphics[width=\textwidth]{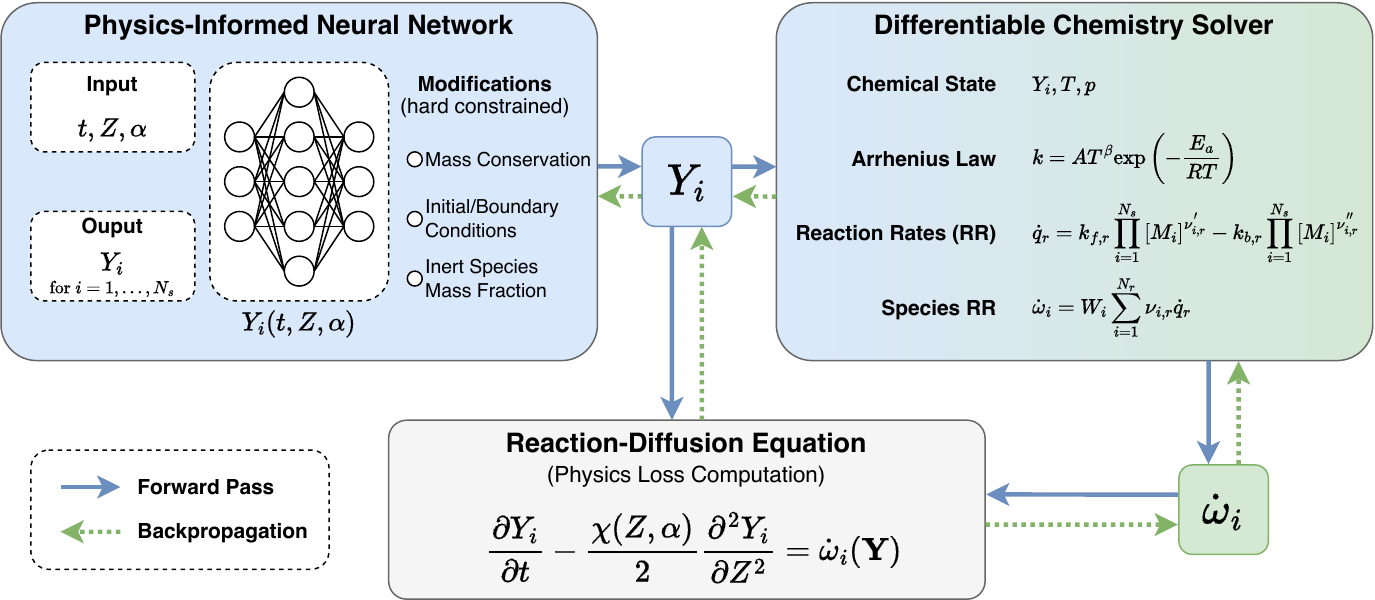}}
    \caption{\textbf{Schematic illustration of the proposed framework.} A modified PINN approximates the species mass fractions $Y_i$. These predictions are evaluated by a differentiable chemistry solver to compute the species reaction rates $\dot\omega_i$. The resulting mass fractions and reaction rates enter the physics-informed loss, enabling end-to-end backpropagation through the chemistry solver during training. For the parameterized problem setup, the strain rate $\alpha$ is taken as an additional input to the network architecture.}
    \label{fig:schematic_drawing}
  \end{center}
\end{figure*}

\section{Methodology}

\subsection{Physics-Informed Neural Networks}
PINNs are a deep learning--based framework for solving forward and inverse problems governed by partial and ordinary differential equations~\cite{RAISSI2019686}. They employ neural networks as universal function approximators \cite{hornik1991approximation} to represent the unknown solution of a differential equation. In the forward setting, full knowledge of the governing equations is used to approximate the solution, while in the inverse setting partial physical knowledge together with observational data is leveraged to infer unknown quantities such as equation parameters.

The governing equation is assumed to be of the form
\begin{equation}
\mathcal{F}(u(x);\mu) = 0, \qquad x \in \Omega,
\label{eq:differential_equation}
\end{equation}
where $\mathcal{F}$ denotes a (nonlinear) differential operator, $u(x)$ is the (vector-valued) solution, $x$ is a multidimensional input variable, $\mu$ is a vector of physical parameters, and $\Omega$ denotes the computational domain.

The solution is approximated by a neural network $u_\theta(x)$ with parameters $\theta$. The corresponding physics residual, capturing the mismatch between the model's physical behavior and what is dictated by Eq.~\eqref{eq:differential_equation}, is defined as
\begin{equation}
r_\theta(x) := \mathcal{F}(u_\theta(x);\mu).
\end{equation}

Training is performed by minimizing a composite loss function of the form
\begin{equation}
\mathcal{L}
=
\mathcal{L}_{\mathrm{f}}
+
\mathcal{L}_{\mathrm{IC/BC}}
+
\mathcal{L}_{\mathrm{data}},
\end{equation}
where the physics loss enforces the governing equations at collocation points sampled from the interior of the domain:
\begin{equation}
\mathcal{L}_{\mathrm{f}}
=
\sum_{i=1}^{N_f}
\left|
r_\theta(x_f^i)
\right|^2,
\qquad x_f^i \in \Omega.
\label{eq:physics_loss}
\end{equation}

The initial and boundary condition loss penalizes deviations from prescribed conditions,
\begin{equation}
\mathcal{L}_{\mathrm{IC/BC}}
=
\sum_{i=1}^{N_u}
\left|
u_\theta(x_u^i) - u(x_u^i)
\right|^2,
\qquad x_u^i \in \partial\Omega,
\end{equation}
which are required to render the problem well posed.

In the presence of observational data, an additional data loss term is included,
\begin{equation}
\mathcal{L}_{\mathrm{data}}
=
\sum_{i=1}^{N_d}
\left|
u_\theta(x_d^i) - \hat{u}(x_d^i)
\right|^2,
\qquad x_d^i \in \Omega,
\end{equation}
where $\hat{u}(x)$ denotes observed data.

In the forward setting, the data loss term is not required provided that sufficient initial and boundary conditions are specified to render the problem well posed. In contrast, in the inverse setting the objective is to infer unknown physical parameters $\mu$ from observed data. In this case, the data loss term is essential, and the parameters $\mu$ are typically treated as additional learnable variables jointly optimized with the network parameters.

\subsection{Reaction Systems and Differentiable Chemistry}

To apply PINNs to reactive (combusting) systems, the governing equation \eqref{eq:differential_equation} is augmented by a chemical source term,

\begin{equation}
\label{eq:differential_equation_react}
\mathcal{F}(Y_i(x);\mu) = \dot{\omega}_i(\boldsymbol{Y}), \qquad x \in \Omega,
\end{equation}

where $Y_i(x)$ denotes the $i$-th component of the species mass fractions vector $\boldsymbol{Y}$. The source term $\dot{\omega}_i$ depends on the full composition and the thermodynamic state, yielding a tightly coupled, nonlinear system across all species. Under constant-pressure conditions ($p=$const), the thermodynamic state is characterized by the temperature $T$ together with $\boldsymbol{Y}$ (or, equivalently, by a prescribed enthalpy $h$ from which $T$ is recovered). Moreover, the kinetics exhibit Arrhenius temperature sensitivity, so that, in mass-action form, 

\begin{equation}
\label{eq:prop_Arrhenius}
\dot{\omega} \propto \exp \left(-\frac{1}{T}\right),
\end{equation}

which contributes strongly to stiffness due to the exponential dependence on $1/T$. Classical numerical simulation discretizes the differential equations in time and in case of PDE, in spatial coordinate. Chemical kinetics are usually solved by suitable chemistry solvers such as cantera~\cite{cantera2025}. Through the source term dependency on the solution of the differential equation, special numerical treatment is required which ranges from linearization to operator splitting schemes in case of reaction-diffusion PDE. Furthermore, the stiff dynamics of reacting systems requires stability constraints leading to limiting step sizes and thus increased simulation times. By contrast, PINNs enables the treatment of the chemistry "instantaneously" at the level of residual satisfaction rather than via sequential time stepping. However, this requires that the solution of the underlying chemistry is part of the computational graph and thus be differentiable. In our proposed PINN, the reaction source terms are determined through the differentiable chemistry backend \textit{reactorch}~\cite{reactorch}. Given the predicted species mass fractions, the mixture enthalpy $h$ is first computed from the boundary stream states, and the corresponding temperature $T$ is recovered implicitly by solving the thermochemical equilibrium relation. The recovered temperature and species composition are then used to evaluate reaction rates via a differentiable implementation of detailed chemical kinetics. This formulation enables end-to-end gradient propagation through the chemistry evaluation and is essential for both forward and inverse problem settings (Figure~\ref{fig:schematic_drawing}).
Further details on the chemical kinetics computations can be found in the appendix in Section~\ref{sec:appendix-chemistry}.

\subsection{Proposed Framework}
 Solving stiff reaction systems requires several modifications to the training procedure, without which stable convergence is not achievable. Due to the extreme stiffness of the governing equations, we incorporate residual weighting as proposed in \cite{rohrhofer2025approximating}. Specifically, each residual term in the physics loss computation (cf. Eq.~\eqref{eq:physics_loss}) is scaled according to the corresponding reaction coefficient,
\begin{equation}
\label{eq:res-weighting}
\rho_i^{(j)} = \frac{1}{1 + \lambda |\omega_i(\boldsymbol{Y}^{(j)})|},
\end{equation}
where $\omega_i(\boldsymbol{Y}^{(j)})$ denotes the reaction rate of species $i$ evaluated for the mass-fraction vector $\boldsymbol{Y}^{j}$, and $\lambda$ is a hyperparameter. This weighting reduces the contribution of regions with very high reaction rates to the loss function, thereby mitigating instabilities commonly encountered in stiff reactive systems.

In the parameterized setting, we do not sample the parameter independently for each collocation point. Instead, for a given parameter value, we sample a full set of collocation points covering the entire spatiotemporal domain. This reflects the fact that each parameter corresponds to a single global solution of the governing equations, which must satisfy the PDE constraints throughout the domain. Sampling collocation points in this structured manner is essential for enforcing global physical consistency and was found to be crucial for stable training in the stiff reactive setting considered here. For the parameterized case, we adopt the network architecture proposed in~\cite{pmlr-v235-cho24b} rather than a standard fully connected architecture.

Initial and boundary conditions are enforced through hard constraints. The inert species (N$_2$) mass fraction is constrained throughout the domain, and the predicted mass fractions of the reactive species are constrained via a softmax transformation to ensure mass conservation. We employ the Adam optimizer for training and use only the physics loss in all forward problem settings. Details of the network architecture and training procedure are provided in Section~\ref{sec:appendix-networks}.

\section{Experimental Setup}
We evaluate the proposed framework on three distinct systems derived from hydrogen combustion chemistry, modeled using a global one-step Arrhenius mechanism to represent the underlying reaction kinetics. The considered problems include an initial value problem, a boundary value problem, and a reaction--diffusion partial differential equation. Hydrogen combustion is used throughout as a deliberately challenging test case due to its extreme stiffness and wide separation of chemical time scales. Rather than varying the chemical mechanism, we focus on evaluating the proposed framework across fundamentally different problem classes (IVPs, BVPs, and reaction–diffusion PDEs), each of which typically requires a distinct numerical treatment. Successful application to hydrogen chemistry thus serves as a strong indicator of robustness and expected generalizability to less stiff combustion systems. In the PDE setting, we study both forward problems, including a parameterized formulation in which a family of solutions indexed by the strain rate is learned, and an inverse problem, where the strain rate is inferred from observed solution data. While each of these systems typically requires a fundamentally different numerical solution strategy, the corresponding PINN formulations differ only minimally, highlighting the generality of the proposed approach.

Since the proposed framework incorporates several training and formulation choices beyond a standard PINN, we evaluate its performance on forward problems against a basic PINN baseline as well as against one well known PINN modification to serve as a competitive baseline. The vanilla PINN baseline enforces only hard initial and boundary conditions and does not include any additional problem-specific constraints, such as mass-fraction normalization or inert-species enforcement, serving as a minimal reference implementation. In addition, we compare against causality-based loss scaling~\cite{wang2022respecting} which is a widely known PINN training modification, well suited for time-dependent systems.

For a fair comparison, all methods share the same network architectures including the hard constraints for initial and boundary conditions and optimization settings. In the case of the competitive baseline, only training modifications that are not orthogonal to the respective method are removed, while all other components of the framework are kept identical. Implementation details are provided in Appendix~\ref{sec:appendix-networks}, and additional experimental details are described in Appendix~\ref{sec:appendix-experiments}.

\subsection{Initial Value Problem}
We first consider an initial value problem arising from hydrogen combustion chemistry, formulated as a coupled system of ordinary differential equations 
\begin{equation}
\frac{d Y_i}{d t} = \omega_i(\boldsymbol{Y}), \qquad i = 1, \dots, N,
\end{equation}
where $\boldsymbol{Y}$ denotes the vector of species mass fractions and $\omega_i(\boldsymbol{Y})$ represents the reaction rate of species $i$ computed through detailed chemical kinetics.

We solve this system using a PINN by enforcing the governing equations at collocation points sampled over the temporal domain. To account for the sharp time derivatives occurring at early times, collocation points are sampled in logarithmic time coordinates, and logarithmic transformation is applied to the input. This setting serves as a baseline test for assessing the ability of the proposed framework to handle stiff reaction dynamics in a purely forward setting.

\subsection{Boundary Value Problem}
Next, we evaluate the proposed framework on a boundary value problem corresponding to the steady-state formulation of the reaction–diffusion PDE describing hydrogen combustion. The system reduces to a second-order ordinary differential equation of the form
\begin{equation}\label{eq:bv-ode}
-\frac{\chi(Z, \alpha)}{2} \frac{d^2 Y_i}{d Z^2} = \omega_i(\boldsymbol{Y}), \qquad i = 1, \dots, N.
\end{equation}
where  \( Z \in [0,1] \) is the mixture fraction and $\frac{\chi(Z, \alpha)}{2}$ is the scalar dissipation rate which is a function of strain rate $\alpha$ and mixture fraction $Z$.
Boundary value problems are known to be particularly challenging for both PINNs and classical numerical solvers.

\subsection{Reaction-Diffusion PDE}

We consider a reaction--diffusion partial differential equation describing hydrogen combustion in a flamelet configuration. The governing equation couples diffusion in mixture--fraction space with nonlinear reaction terms arising from detailed chemical kinetics and is given by
\begin{equation}\label{eq:pde-flamelet}
\frac{\partial Y_i}{\partial t}
- \frac{\chi (Z, \alpha)}{2}\frac{\partial^2 Y_i}{\partial Z^2}
= \omega_i(\boldsymbol{Y}), 
\qquad i = 1, \dots, N,
\end{equation}

 In the forward setting, the strain rate is fixed and the network is trained to approximate the corresponding flamelet solution for a given parameter value. This fixed-parameter formulation is used as the primary benchmark for performing the ablation study. In the inverse setting, the strain rate is treated as an unknown parameter and is inferred by simultaneously minimizing the physics-informed loss and the data loss.

Beyond these fixed-parameter experiments, we additionally consider a parameterized formulation in which the strain rate is provided as an explicit input to the network. This enables learning a continuous family of flamelet solutions across a prescribed range of strain rates using a single neural network. The parameterized setting serves as an extension of the core methodology and illustrates the flexibility of the proposed framework in handling forward problem across parameter space.

\section{Evaluation and Results}
In Table \ref{table:forward-results} we show the quantitative results for all of the forward problem settings we considered in our evaluation. Namely, we compare our proposed framework with regular PINN training (without any chemistry specific modifications), and also with causality-based loss scaling approach as a competitive baseline. We compute the mean absolute error and also the relative $L_2$ error on initial value problem, boundary value problem, reaction-diffusion PDE in both non-parametrized setting with a fixed strain rate and with the parametrized setting with strain rate used as input. Our results show that the PINN with our proposed modifications outperforms both baselines across all the test cases.

\begin{table*}[t]
  \caption{Forward problem results for all tested systems. We report mean absolute error (MAE) and relative $L_2$ error with respect to numerical reference solutions. For the parameterized PDE, metrics are aggregated over strain rates $\alpha \in [1,100]$.}
  \label{table:forward-results}
  \begin{center}
    \begin{small}
      \begin{sc}
        \begin{tabular}{lcccccc}
          \toprule
          \multirow{2}{*}{System} 
          & \multicolumn{2}{c}{Vanilla PINN}
          & \multicolumn{2}{c}{Causality-based PINN}
          & \multicolumn{2}{c}{Proposed framework} \\
          \cmidrule(lr){2-3}\cmidrule(lr){4-5}\cmidrule(lr){6-7}
          & MAE & Rel.\,$L_2$
          & MAE & Rel.\,$L_2$
          & MAE & Rel.\,$L_2$ \\
          \midrule
          IVP ODE
          & 6.60e-01 & 3.81e-00
          & 7.91e-03 & 4.80e-02
          & 1.10e-03 & 4.37e-03 \\

          BVP ODE ($\alpha{=}1$)
          & 2.17e-00 & 7.27e-00
          & -- & --
          & 2.80e-03 & 1.71e-02 \\

          BVP ODE ($\alpha{=}100$)
          & 2.03e-00 & 6.79e-00
          & -- & --
          & 5.20e-03 & 2.26e-02 \\

          PDE ($\alpha{=}1$)
          & 8.81e-02 & 7.04e-01
          & 4.03e-03 & 2.62e-02
          & 2.41e-04 & 1.90e-03 \\

          PDE ($\alpha{=}100$)
          & 7.17e-02 & 5.19e-01
          & 3.32e-03 & 2.22e-02
          & 6.59e-05 & 3.79e-04 \\

          Parametrized PDE ($\alpha \in [1,100]$)
          & 6.33e-03 & 4.10e-02
          & 2.40e-03 & 1.52e-02
          & 8.74e-04 & 7.63e-03 \\
          \bottomrule
        \end{tabular}
      \end{sc}
    \end{small}
  \end{center}
  \vskip -0.1in
\end{table*}

\subsection{Initial Value ODE}
\begin{figure}[ht]
  \vskip 0.2in
  \begin{center}
    \centerline{\includegraphics[width=\columnwidth]{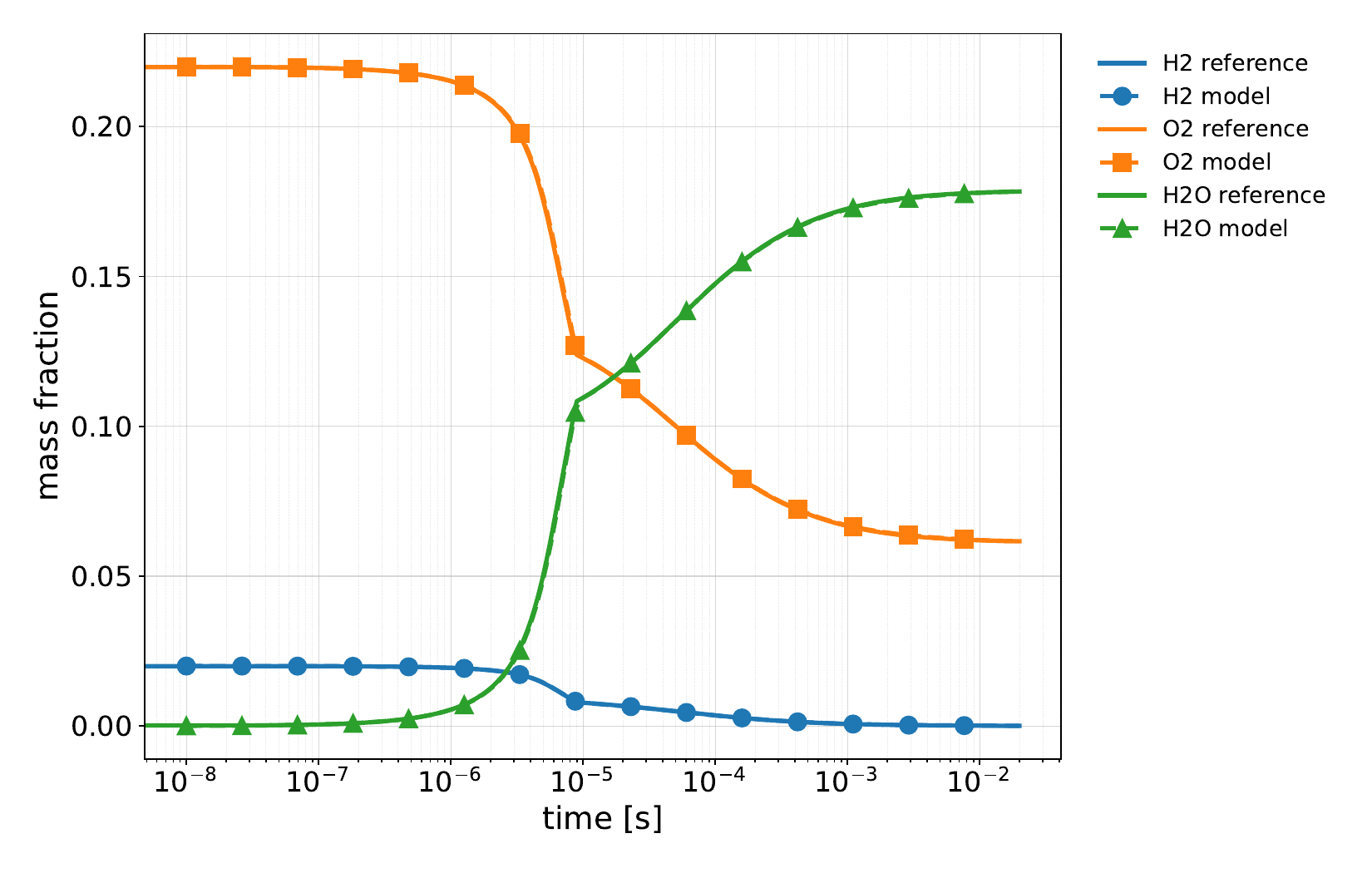}}
    \caption{
      Results for initial value problem ODE.
    }
    \label{fig:ODE-forward}
  \end{center}
\end{figure}

Figure \ref{fig:ODE-forward} shows the solution of the stiff initial value problem obtained using the proposed PINN. Despite the presence of rapid transients at early times, the network accurately captures the evolution of all species across the full temporal domain. This experiment demonstrates that the proposed framework can robustly handle highly stiff reaction dynamics in a purely forward setting.

\subsection{Boundary Value ODE}
\begin{figure}[ht]
  \vskip 0.2in
  \begin{center}
     \centerline{\includegraphics[width=\columnwidth]{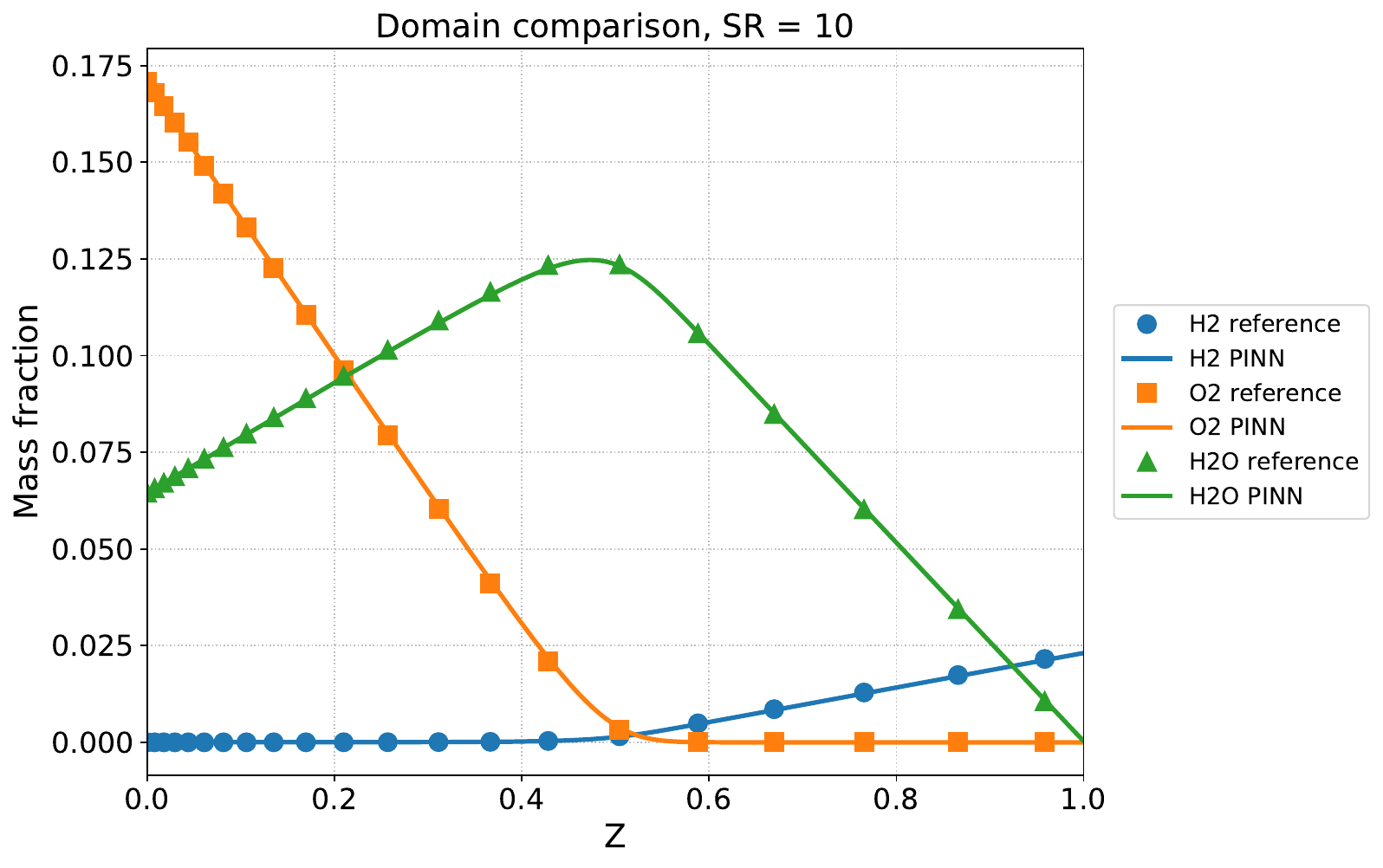}}
    \caption{
      Results for boundary value problem ODE with the strain rate $\alpha$ set to 10.
    }
    \label{fig:ODE-boundary}
  \end{center}
\end{figure}

Figure~\ref{fig:ODE-boundary} presents the solution of the boundary value problem ODE corresponding to the steady-state solution of the PDE flamelet formulation. We note that although this problem does not exhibit sharp temporal transients and the solution appears relatively smooth, training without residual scaling led to severe convergence issues and inaccurate results (not shown).

\subsection{Reaction-Diffusion PDE}
\subsubsection{Forward Problem}
\begin{figure}[ht]
  \vskip 0.2in
  \begin{center}
    \centerline{\includegraphics[width=\columnwidth]{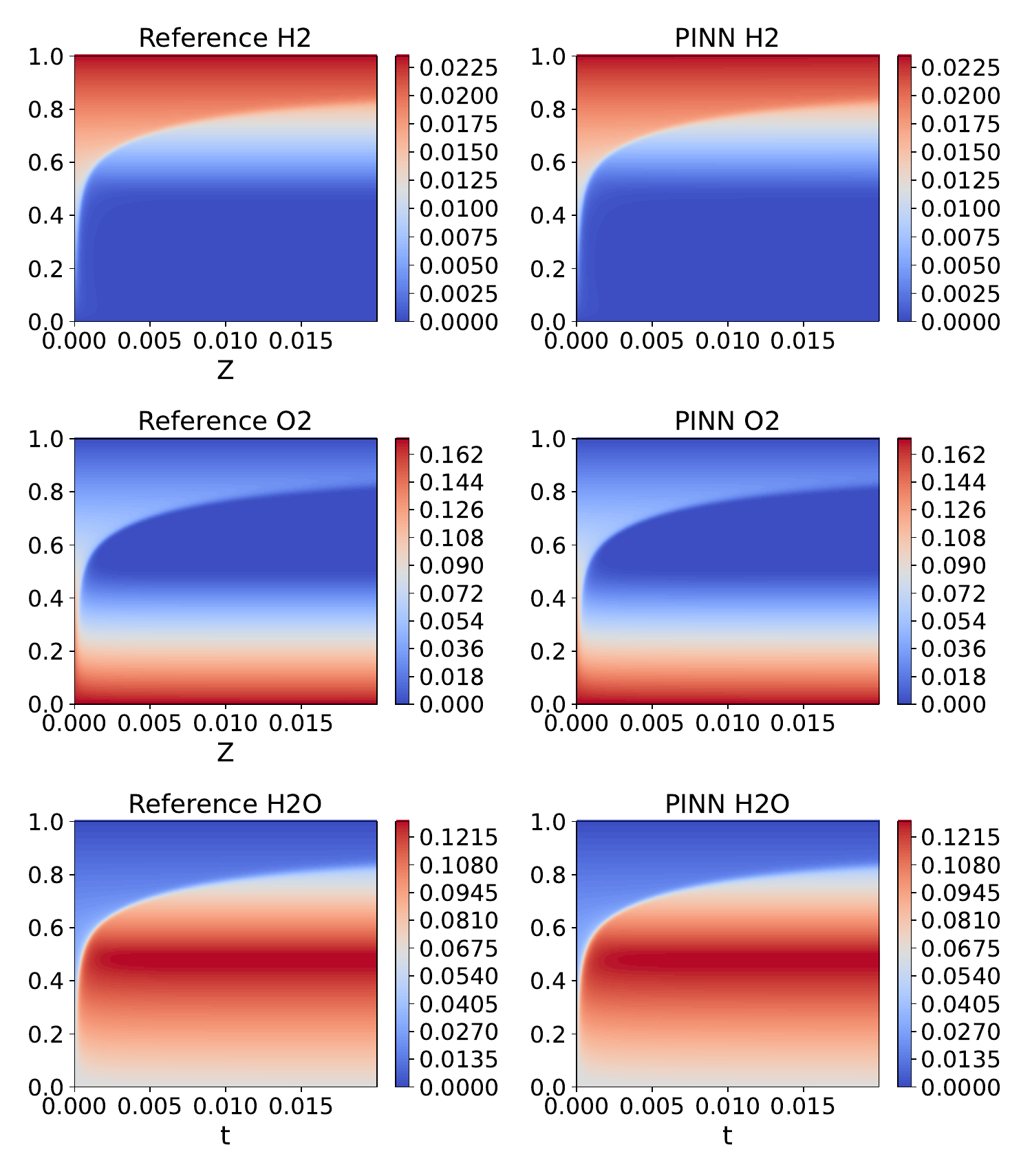}}
    \caption{
      Results for flamelet PDE PINN for strain rate $\alpha=1$. The left columns shows the reference species mass fraction, the right column the PINN predictions using the proposed framework.
    }
    \label{fig:PDE-SR1}
  \end{center}
\end{figure}

Figure~\ref{fig:PDE-SR1} shows the solution of the reaction-diffusion flamelet PDE for a fixed strain rate. The proposed PINN accurately reproduces the species evolution across the full spatiotemporal domain without exhibiting spurious oscillations or instability. Importantly, stable convergence is achieved in a regime characterized by extremely stiff and sharply varying reaction dynamics, which is widely regarded as challenging for standard PINN formulations. To the best of our knowledge, this constitutes the first successful application of PINNs to reaction-diffusion PDE with detailed chemistry. This experiment therefore provides a representative benchmark for evaluating stabilization strategies in stiff reactive PDEs. 

\subsubsection{Inverse Problem}
Table~\ref{table:inverse} reports the results of the inverse problem for the reaction-diffusion flamelet PDE, where the strain rate~$\alpha$ is inferred from observed solution data. The evaluation is performed for three representative ground-truth values, $\alpha\in\{1,10,100\}$, and different additive noise levels applied to the observational data.

Across all cases, the PINN accurately recovers the true strain rate, even in the cases with large levels of additive noise. These results demonstrate that the proposed framework is able to identify global physical parameters directly from the PDE residual, even in the presence of stiff reaction dynamics and highly nonlinear source terms.

\begin{table}[t]
  \caption{Inverse problem: inferred strain rate $\hat{\alpha}$ for different true values $\alpha$ under additive noise in the observations.}
  \label{table:inverse}
  \begin{center}
    \begin{small}
      \begin{sc}
        \begin{tabular}{lccc}
          \toprule
          \multirow{2}{*}{Noise level}
          & \multicolumn{3}{c}{True strain rate $\alpha$} \\
          \cmidrule(lr){2-4}
          & 1.0 & 10.0 & 100.0 \\
          \midrule
          0.0  & 0.994  & 10.002 & 99.624 \\
          0.01 & 0.995  & 9.992     & 99.598     \\
          0.1  & 0.998  & 10.022  & 99.281  \\
          1.0 & 1.027 & 10.199 & 94.657 \\
          \bottomrule
        \end{tabular}
      \end{sc}
    \end{small}
  \end{center}
  \vskip -0.1in
\end{table}

\subsubsection{Parametrized Problem}
\begin{figure}[ht]
  \vskip 0.2in
  \begin{center}
    \centerline{\includegraphics[width=\columnwidth]{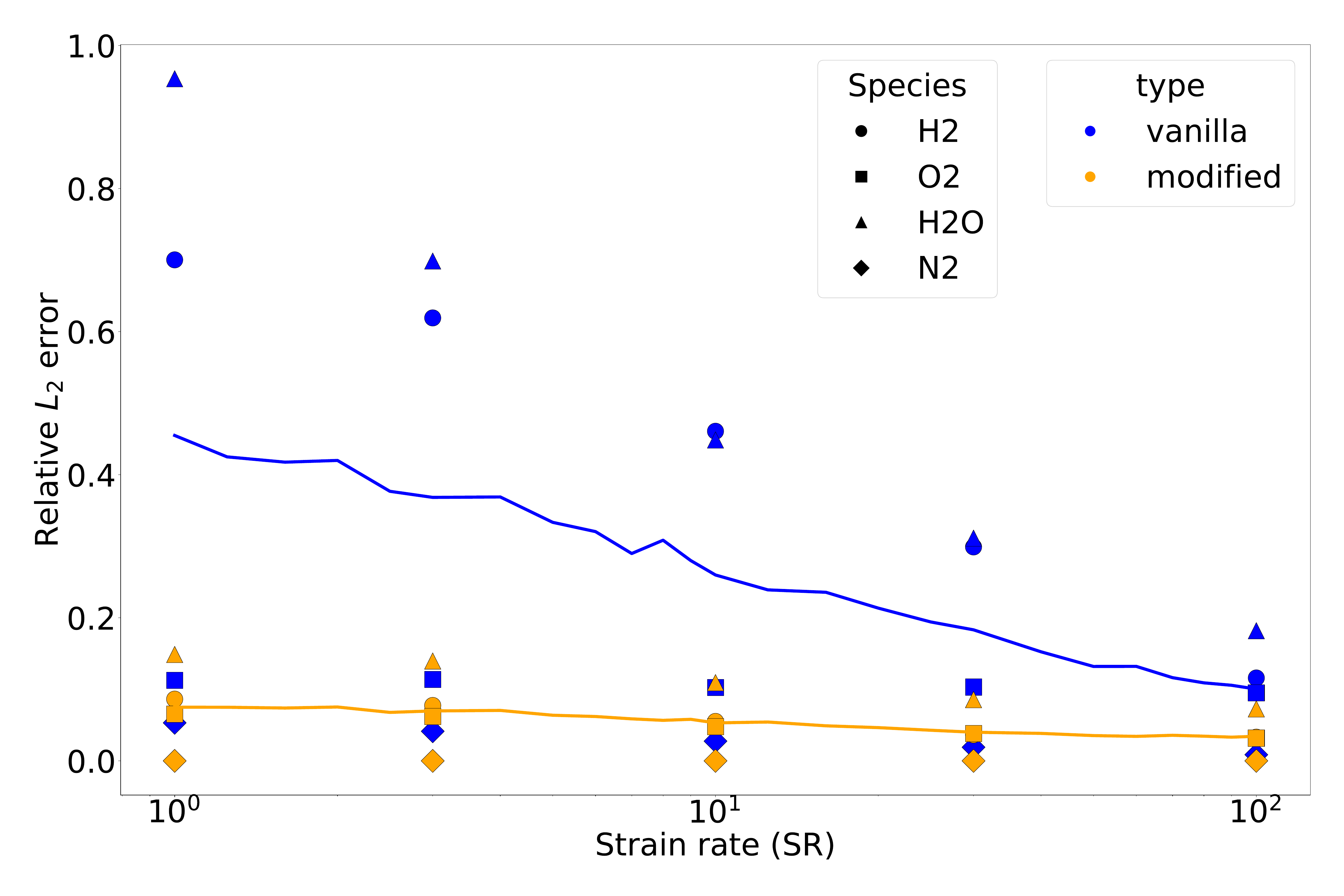}}
    \caption{
      Relative $L_2$ error resolved by individual species and strain rates comparing vanilla PINN approach without domain specific modifications and our proposed framework with multiple modifications that take stiffness of the system into account. Lines represent relative $L_2$ error averaged over all species.
    }
    \label{fig:parametrized-comparison}
  \end{center}
\end{figure}

Figure~\ref{fig:parametrized-comparison} compares the performance of the parameterized PINN against a vanilla PINN baseline across a range of strain rates. The error is resolved both by strain rate and by individual chemical species. Two trends are immediately apparent. First, lower strain rates are consistently more difficult for PINNs to learn, which is consistent with the increased stiffness of the underlying reaction dynamics in this regime. Second, the relative $L_2$ error is notably higher for hydrogen compared to other species. This can be attributed to its very low mass fraction over large parts of the domain, which leads to an imbalance in the residual contributions and makes accurate learning more challenging for standard PINN formulations.

In contrast, the proposed framework significantly reduces the error across all strain rates and species. In particular, the improvement is most pronounced at low strain rates, where stiffness effects are strongest and vanilla PINNs struggle to converge to physically meaningful solutions. 

\begin{figure}[ht]
  \vskip 0.2in
  \begin{center}
    \centerline{\includegraphics[width=\columnwidth]{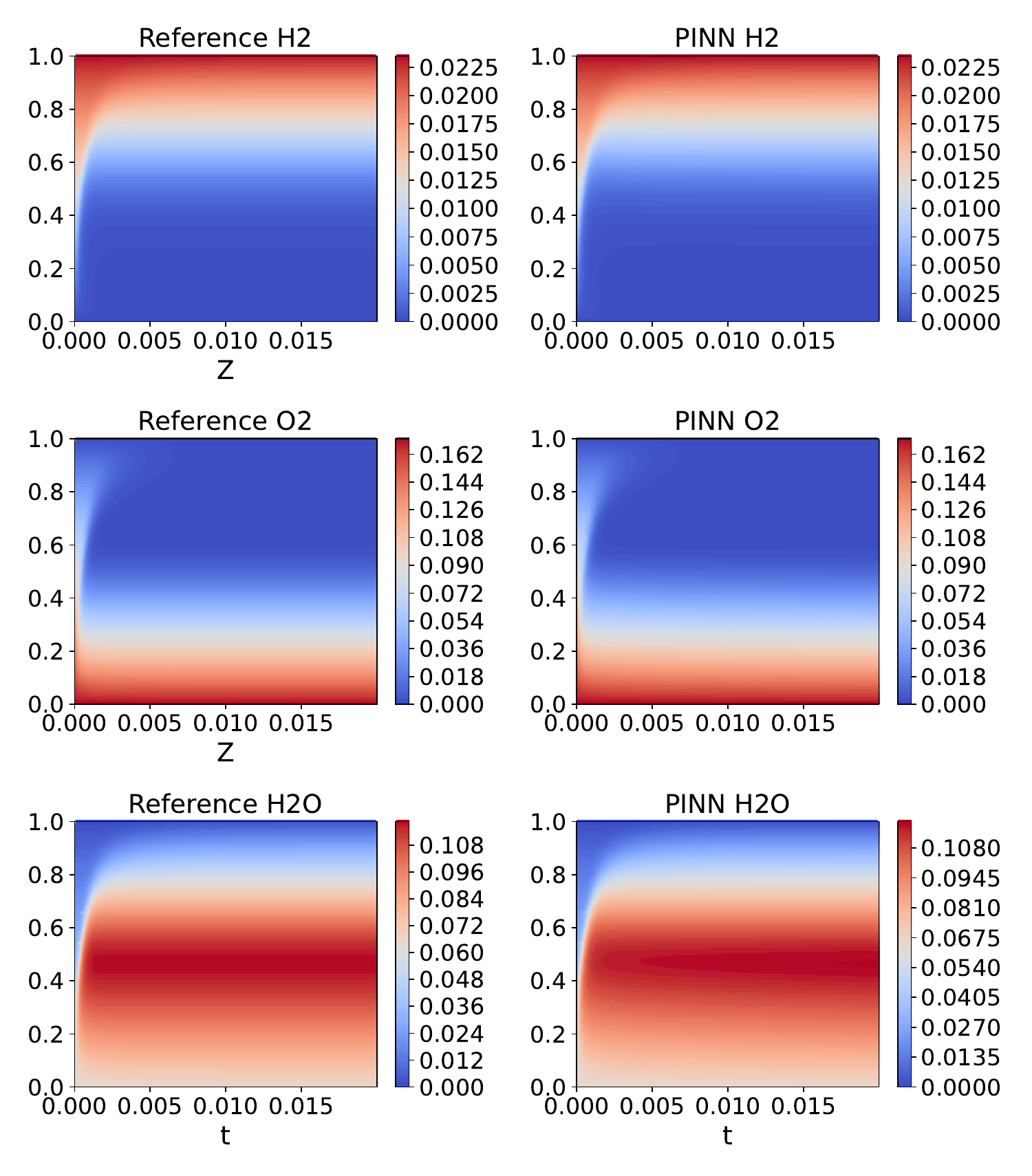}}
    \caption{
      Results for the parametrized PINN case trained on $\alpha \in [1, 100]$ ploted for the $\alpha=100$.
    }
    \label{fig:parametrized-PDE-100}
  \end{center}
\end{figure}

To further illustrate the quality of the learned parameterized representation, Figure~\ref{fig:parametrized-PDE-100} shows the predicted flamelet solution evaluated at a representative strain rate of $\alpha = 100$. The parameterized PINN accurately reproduces the spatiotemporal structure of the solution, demonstrating that a single network can represent a continuous family of flamelet solutions.

In practical combustion simulations, flamelet-based models often rely on precomputed solution tables spanning a range of strain rates and additional physical parameters. While tabulation is effective in low-dimensional settings, the size of such tables can grow rapidly as more parameters are introduced, leading to increased memory requirements and interpolation complexity. Although the present experiments consider a regime where tabulation remains practical, the parameterized PINN formulation explored here suggests a potential path toward memory efficient representation of the flamelet equation solutions. With further development, such neural representations could enable simulations that would otherwise be impractical due to prohibitive memory requirements.

\subsubsection{Ablation Study}
For the forward reaction-diffusion PDE, we perform an ablation study to assess the contribution of chemistry-related modifications introduced in the proposed framework. Starting from the full model, we selectively remove individual components while keeping all other training settings fixed.

Specifically, we examine the impact of removing (i) differentiability of the chemistry solver, (ii) residual scaling of the PDE loss, (iii) the hard constraint on the inert species mass fraction, and (iv) the mass conservation constraint. 

Table~\ref{table:ablation} reports the mean absolute error (MAE) and relative $L_2$ error for the full model and the ablated variants. Removing any of the proposed components leads to a significant degradation of accuracy.

\begin{table}[t]
  \caption{Ablation study on the reaction-diffusion PDE in the forward setting with $\alpha=1$.}
  \label{table:ablation}
  \begin{center}
    \begin{small}
      \begin{sc}
        \begin{tabular}{lccr}
          \toprule
          Method  &       MAE     & Rel. $L_2$  \\
          \midrule
          Proposed PINN     & 2.41e-04 & 1.90e-03 \\  
          Non-diff. chem. solver    & 6.77e-03 & 3.26e-02 \\
          No residual scaling   & 2.85e-03 & 1.71e-02 \\
          No inert specie constraint & 2.50e-03 & 1.65e-02 \\
          No mass conservation & 3.19e-03 & 2.03e-02 \\
          \bottomrule
        \end{tabular}
      \end{sc}
    \end{small}
  \end{center}
  \vskip -0.1in
\end{table}

\section{Conclusion}
In this work, we proposed a general framework for integrating differentiable chemistry solvers with physics-informed neural networks to solve combustion-related differential equations involving detailed chemical kinetics. A recent comprehensive review~\cite{wu2025} confirms that, to the best of our knowledge, this represents the first successful demonstration of PINNs applied to time-dependent reaction-diffusion PDEs involving physically grounded chemical kinetics in a purely forward setting using physics-based losses only. 

We further demonstrated that the proposed framework can be extended to a parameterized formulation, enabling a single neural network to approximate a continuous family of solutions across a wide range of strain rates. This capability suggests a potential pathway toward integrating PINN-based chemistry models into high-fidelity CFD simulations, where detailed chemical kinetics are traditionally represented through discrete precomputed lookup tables. Such neural representations could offer a more flexible and memory-efficient alternative for modeling combustion chemistry across parameter spaces.

The present study is limited to global one-step Arrhenius reaction mechanisms. While the proposed framework is, in principle, directly applicable to multi-step and detailed chemical kinetics, extending it to such settings introduces significant additional challenges for PINN optimization due to increased stiffness and system dimensionality. Addressing these challenges remains an important direction for future work.

\section*{Acknowledgment}
The financial support by the Austrian Federal Ministry of Economy, Energy and Tourism, and the Christian Doppler Research Association is gratefully acknowledged. Furthermore, the authors want to thank Prof. Ricardo Novella and Prof. José Maria García-Oliver from the Clean Mobility \& Thermofluids department of the Universitat Politècnica de València for their input on flamelet data generation.

The work Franz M. Rohrhofer was partially funded by the European Union’s HORIZON Research and Innovation Programme under grant agreement No 101120657, project ENFIELD (European Lighthouse to Manifest Trustworthy and Green AI). Know Center is a COMET competence center that is financed by the Austrian Federal Ministry of Innovation, Mobility and Infrastructure (BMIMI), the Austrian Federal Ministry of Economy, Energy and Tourism (BMWET), the Province of Styria, the Steirische Wirtschaftsförderungsgesellschaft m.b.H. (SFG), the Vienna business agency and the Standortagentur Tirol . The COMET programme is managed by the Austrian Research Promotion Agency FFG.
\FloatBarrier
\bibliography{example_paper}
\bibliographystyle{icml2025}

\newpage
\appendix
\onecolumn
\section{Chemistry}\label{sec:appendix-chemistry}
The underlying chemistry of the reacting system can be described as follows. Let $N_r$ be the number of reactions and $N_s$ the number of species, a general reaction takes the form

\begin{equation}
    \sum_{i=1}^{N_s} \nu'_{i,r}M_i \ch{<=>} \sum_{i=1}^{N_s} \nu''_{i,r}M_i, \hspace{1cm} r=1,...,N_r,
\end{equation}

where $\nu'_{i,r}$ and $\nu''_{i,r}$ are the molar stoichiometric coefficients with

\begin{equation}
    \nu_{i,r}=\nu''_{i,r}-\nu'_{i,r}.
\end{equation}

The reaction rate $\dot{q}_r$ at which each reaction progresses is the difference of the forward and backward reaction rates as

\begin{equation}
    \dot{q}_r=k_{f,r} \prod_{i=1}^{N_s} \left[M_i\right]^{\nu'_{i,r}}- k_{b,r}\prod_{i=1}^{N_s} \left[M_i\right]^{\nu''_{i,r}},
\end{equation}

where $k_{f,r}$ and $k_{b,r}$ denote the forward and backward reaction rates, respectively. The specific reaction rates are determined by the Arrhenius law

\begin{equation}
\label{eq:Arrhenius}
    k=AT^\beta \exp{\left(-\frac{E_a}{RT}\right)},
\end{equation}

where $A$ and $\beta$ are constants, $E_a$ is the activation energy, $R$ is the universal gas constant, and $T$ is the temperature. From this law, the strong influence of the temperature on the reaction can be seen. At low temperatures, the reaction velocity is very low while with increasing temperature, reactions become extremely fast leading to the considerable stiffness of the formalism to be solved. The species reaction rate $\dot{\omega}_i$ is calculated by

\begin{equation}
    \dot{\omega}_i = W_i \sum_{r=1}^{N_r} \nu_{i,r}\dot{q}_r,
\end{equation}

with $W_i$ being the molecular weight of species $i$.

Additionally to the chemical kinetics, the modeling of the thermodynamic properties of the desired species is required. This is utilized by applying the NASA 7-coefficient polynomial parameterization to compute the species standard state thermodynamic properties $\hat{c}^0_p\left(T\right)$, $\hat{h}^0\left(T\right)$, and $\hat{s}^0\left(T\right)$.

In the present study, we focused on hydrogen combustion, since its fast kinetics represents major challenges for the PINN in solving the resulting stiffness of the problem. We model the chemistry with a single-step global reaction 

\begin{equation}
    2 \ch{H2} + \ch{O2} \ch{<=>} 2\ch{H2O},
\end{equation}

with the Arrhenius constants, cf.(\ref{eq:Arrhenius}), $A=1.8\mathrm{e}19$, $\beta=0$ and $E_a=17614$. The thermodynamic properties are taken from the well-known San Diego hydrogen mechanism~\cite{Saxena2006}.

In order to use the differentiable solver \textit{reactorch} within our boundary-value and reaction–diffusion formulations, we extended the library to support an enthalpy-based thermodynamic state. The original API specifies the state via pressure $p$, temperature $T$, and species mass fractions $\boldsymbol{y}$, whereas our formulation prescribes mixture enthalpy as a function of mixture fraction and the enthaply values of the fuel and oxidizer, respectively, as 

\begin{equation}
    h = Z\cdot h_{\mathrm{fuel}}+\left(1-Z\right)\cdot h_{\mathrm{oxidizer}}.
\end{equation}

We therefore added an interface that accepts $\left(p,h,\boldsymbol{y}\right)$ using a root finder for solving the NASA 7-coefficient polynomial. Of course, this root finder has to be differentiable to maintain the end-to-end gradient propagation.

\section{Reference Data Generation}
Across all cases, the reference data are produced with classical numerical solvers. Tolerances and resolutions are chosen such that the remaining discretization/solver error is negligible compared to the reported PINN errors.

\subsection{Initial Value Problem}
The reference trajectories are generated with SciPy’s \textit{solve\_ivp} integrator. We use an implicit method (Radau) to resolve the stiff behavior of the hydrogen reaction. Absolute and relative tolerances are set to stringent values (atol=1e-12, rtol=1e-8) and the solution is evaluated on a dense temporal grid.

\subsection{Boundary Value Problem and Reaction-Diffusion PDE}
For generating the reference data to validate the boundary value problem and the reaction-diffusion PDE, we used the flamelet solver \textit{ZLFLAM}~\cite{NAUD2015}. The solver uses Chemkin to calculate the reaction source terms $\dot\omega_i\left(\boldsymbol{Y}\right)$ based on the specified chemical mechanism. Furthermore, the "Twopnt program for boundary value problems" presented in~\cite{grcar1991} was used for solving the boundary value problem in $Z$. The unsteady solution is performed applying the DDASSL solver~\cite{petzold1982} using the exact block tridiagonal Jacobian matrix. For the discretization of $Z$, we use a non-conformal resolution 201 in the boundary value problem and the reaction-diffusion PDE.

The scalar dissipation rate $\chi$, denoting the diffusion coefficient in \eqref{eq:bv-ode} and \eqref{eq:pde-flamelet} is calculated according to~\cite{Peters2000} by

\begin{equation}
   \chi\left(Z,\alpha\right)=\frac{\alpha}{\pi}\exp\left[-2\left(\mathrm{erc}^{-1} \left(2Z\right)\right)^2 \right],
\end{equation}

with $\mathrm{erfc}\left(x\right)=1-\mathrm{erf}\left(x\right) = \left(2/\sqrt{\pi}\right)\int_x^\infty \exp\left(-y^2\right)\mathrm{d}y$.

\section{Network Architecture and Training Settings}\label{sec:appendix-networks}
\subsection{Initial Value Problem}\label{sec:ivp-setup}

The solution for the initial value problem is approximated using a fully connected neural network with four hidden layers and 50 neurons per layer. All hidden layers use the hyperbolic tangent activation function, while an exponential activation is applied to the output layer to ensure non-negativity of the predicted mass fractions and also to allow applying constraint for the mass fraction sum. The network is trained using the Adam optimizer with a fixed learning rate of $10^{-3}$ for 20{,}000 epochs.

To improve numerical stability in the presence of sharp temporal transients at early times, the temporal input is transformed to logarithmic scale. Specifically, the network takes as input $\log(t + \varepsilon)$, where $\varepsilon > 0$ is a small constant to avoid singularities at $t=0$. This transformation allows the network to better resolve the stiff initial dynamics commonly encountered in reactive systems.

Hard constraints are imposed to enforce the initial conditions, mass conservation, and the constancy of the inert species mass fraction. Let $\boldsymbol{y}(t) \in \mathbb{R}^N$ denote the vector of species mass fractions, and let $y_{\mathrm{N_2}}$ denote the inert nitrogen mass fraction. The network predicts only the reactive species mass fractions through an unconstrained neural output $\boldsymbol{f}_\theta(\log t + \varepsilon) \in \mathbb{R}^{N-1}$. The reactive species mass fractions are constructed as
\begin{equation}
\tilde{\boldsymbol{Y}}_{\mathrm{react}}(t)
=
\boldsymbol{Y}_{\mathrm{react}}(0)
+
t \, \boldsymbol{f}_\theta(\log t + \varepsilon),
\end{equation}
which enforces the initial condition exactly at $t=0$. The inert species mass fraction is fixed as
\begin{equation}
Y_{\mathrm{N_2}}(t) = Y_{\mathrm{N_2}}(0),
\end{equation}
and mass conservation is enforced by normalizing the reactive species such that
\begin{equation}\label{eq:mass-conservation}
\boldsymbol{Y}_{\mathrm{react}}(t)
=
\frac{\tilde{\boldsymbol{Y}}_{\mathrm{react}}(t)}
{\sum_i \tilde{Y}_{\mathrm{react},i}(t)}
\left(1 - Y_{\mathrm{N_2}}(0)\right).
\end{equation}
The full mass-fraction vector is then obtained by concatenating the reactive and inert species. This construction guarantees that all predicted mass fractions are non-negative, satisfy the prescribed initial conditions, and sum to $1$ for all $t$.

Physics-informed training is performed by minimizing the mean squared residual of the governing ODE evaluated at collocation points sampled logarithmically in time. Specifically, we draw samples $\xi_j \sim \mathcal{U}(0,1)$ and map them to physical time as
\begin{equation}
t_j
=
\exp\!\left(
\log(t_{\min} + \varepsilon)
+
\xi_j \bigl(\log(t_{\max}) - \log(t_{\min} + \varepsilon)\bigr)
\right),
\qquad j = 1, \dots, N_{\mathrm{col}},
\end{equation}
where $[t_{\min}, t_{\max}]$ denotes the temporal domain, $N_{\mathrm{col}}$ is the number of collocation points, and $\varepsilon > 0$ is a small constant to avoid numerical singularities. This sampling strategy concentrates collocation points near early times while still covering the full temporal domain, enabling accurate enforcement of the physics residual in regions with sharp gradients. Number of collocation points at each epoch is set to $1024$.

\subsection{Boundary Value Problem}\label{sec:bvp-setup}

For solving the boundary value problem we again use the fully connected architecture with 4 hidden layers of 50 units each, hyperbolic tangent activation function in the hidden layers and exponential function in the output layer. In this case we use only soft constraints for the boundary conditions as it, counter-intuitively, improved the training. We again train using the Adam optimizer with a fixed learning rate $0.001$ and train the model for $20,000$ epochs.

We again use hard constraints for the inert species by defining the mass fraction as
\begin{equation}\label{eq:inert-bv}
Y_{\mathrm{N_2}}(t) = Z Y_{fuelN_2} +  (1 - Z) Y_{coflow N_2}.
\end{equation}
We use the same constraint outlined in \eqref{eq:mass-conservation} for mass conservation.

To improve conditioning, the input mixture fraction is scaled to $[-1,1]$ via
\begin{equation}
\hat{Z} = 2\frac{Z - Z_{\min}}{Z_{\max}-Z_{\min}} - 1,
\end{equation}
and the network prediction is evaluated as $\boldsymbol{y}_\theta(Z) = f_\theta(\hat{Z})$.

We enforce the ODE residual at collocation points sampled uniformly in the domain. Specifically, we draw $u_j \sim \mathcal{U}(0,1)$ and map to physical mixture fraction via
\begin{equation}
Z_j = Z_{\min} + (Z_{\max}-Z_{\min})u_j, \qquad j=1,\dots,N_{\mathrm{col}}.
\end{equation}
Derivatives with respect to $Z$ are obtained through automatic differentiation.

Dirichlet boundary conditions are enforced at $Z=0$ and $Z=1$ using the known stream compositions (coflow and fuel states). Let $\boldsymbol{y}_{\mathrm{coflow}}$ and $\boldsymbol{y}_{\mathrm{fuel}}$ denote the corresponding mass-fraction vectors. We include a boundary loss term
\begin{equation}
\mathcal{L}_{\mathrm{BC}}
=
\left\|\boldsymbol{y}_\theta(0)-\boldsymbol{y}_{\mathrm{coflow}}\right\|_2^2
+
\left\|\boldsymbol{y}_\theta(1)-\boldsymbol{y}_{\mathrm{fuel}}\right\|_2^2.
\end{equation}
In addition, we impose soft Neumann-type constraints motivated by the physical boundary behavior of the major reactants. Specifically, we penalize the derivatives
\begin{equation}
\left.\frac{\partial y_{\mathrm{H_2}}}{\partial Z}\right|_{Z=0} \approx 0,
\qquad
\left.\frac{\partial y_{\mathrm{O_2}}}{\partial Z}\right|_{Z=1} \approx 0,
\end{equation}
through the auxiliary loss
\begin{equation}
\mathcal{L}_{\partial Z}
=
\left\|\left.\frac{\partial y_{\mathrm{H_2},\theta}}{\partial Z}\right|_{Z=0}\right\|_2^2
+
\left\|\left.\frac{\partial y_{\mathrm{O_2},\theta}}{\partial Z}\right|_{Z=1}\right\|_2^2,
\end{equation}
where the derivatives are computed by automatic differentiation of the network output with respect to $Z$.

The physics-informed loss consists of the mean squared ODE residual evaluated at collocation points, together with boundary penalties:
\begin{equation}
\mathcal{L}
=
\lambda_{\mathrm{f}}\mathcal{L}_{\mathrm{f}}
+
\mathcal{L}_{\mathrm{BC}}
+
\mathcal{L}_{\partial Z},
\end{equation}
where $\lambda_{\mathrm{f}}$ balances the residual and boundary terms (in our implementation we use a fixed scaling factor of $0.1$).

\subsection{Reaction-Diffusion PDE}
In all cases, the input variables are scaled to the interval $[-1,1]$. Collocation points in time ($t$) and mixture fraction ($Z$) are sampled using Latin hypercube sampling, with $1024$ collocation points per epoch.

Hard constraints for PDE PINN are implemented as
\begin{equation}
    \tilde{\boldsymbol{Y}}_{\mathrm{react}}(t, Z)
=
Z \boldsymbol{Y}_{fuel} + (1 - Z) \boldsymbol{Y}_{coflow} + \boldsymbol{f}_\theta(t, Z) (t / t_{max}) Z (1 - Z).
\end{equation}

For both the forward and inverse problems with a fixed strain rate, we use the same network architecture: a fully connected neural network with four hidden layers of 50 units each, hyperbolic tangent activation functions in the hidden layers, and an exponential activation function at the output. In both cases, training is performed using the Adam optimizer with a fixed learning rate of $0.0005$ for $50{,}000$ epochs. The inert species mass fraction is defined in the same manner as in the boundary value problem, as shown in \eqref{eq:inert-bv}, and mass conservation is enforced as described in \eqref{eq:mass-conservation}. The only difference between the two settings is that the forward problem is trained solely using the physics loss, minimizing the scaled PDE residuals, whereas in the inverse setting the loss function is defined as
\begin{equation}
\mathcal{L} =
\lambda_{\mathrm{f}}\mathcal{L}_{\mathrm{f}}
+
\mathcal{L}_{\mathrm{data}},
\end{equation}
where the loss scaling factor $\lambda_{\mathrm{f}}$ is set to $0.001$. In the inverse setting, the logarithm of the strain rate is treated as a learnable parameter and is used for computing the physics loss.

For the parameterized case, the strain rate is included explicitly as an input to the network. As outlined in the main part of the paper, a single strain rate is sampled per epoch and concatenated with the sampled collocation points for $t$ and $Z$. We employ a specialized architecture for the parameterized case as described in \cite{pmlr-v235-cho24b}. This architecture uses a branched network structure, in which the parameter input ($\alpha$) and the coordinate inputs ($t$, $Z$) are processed by separate encoder networks. The encoded outputs are then concatenated and passed through a decoder network. In our implementation, both encoder networks and the decoder consist of two hidden layers with 50 units each.

\section{Experiment setup}\label{sec:appendix-experiments}

\subsection{Systems Settings}
For all of the systems we used in our experiments, we need to define pressure, which we set to normal atmospheric pressure of $101325.0$ Pa. In the case of initial value problem we set the initial temperature to $1000$ K  and initial mass fractions to:
\begin{align*}
    Y_{H2}(0) &= 0.02 \\
    Y_{O2}(0) &= 0.22 \\
    Y_{H2O}(0) &= 0.0 \\
    Y_{N2}(0) &= 0.76.
\end{align*}
For both the boundary value problem and the reaction-diffusion PDE, the flamelet formulation is defined by two opposing streams corresponding to the fuel and coflow (oxidizer) states. The fuel and coflow states are specified by their temperature and composition according to~\cite{NAUD2015} as
\begin{align*}
\text{Fuel:} \quad
& T = 305~\mathrm{K}, \quad
\boldsymbol{Y}_{\mathrm{fuel}} =
\{ Y_{\mathrm{H_2}} = 0.25,\;
   Y_{\mathrm{N_2}} = 0.75 \}, \\
\text{Coflow:} \quad
& T = 1045~\mathrm{K}, \quad
\boldsymbol{Y}_{\mathrm{coflow}} =
\{ Y_{\mathrm{O_2}} = 0.14744,\;
   Y_{\mathrm{N_2}} = 0.75363,\;
   Y_{\mathrm{H_2O}} = 0.09893 \}.
\end{align*}
Using these definitions, the exact thermochemical states of the fuel and coflow streams are computed with \textit{Cantera}. The resulting species mass fractions define Dirichlet boundary conditions in mixture-fraction space,
\begin{equation}
\boldsymbol{Y}(Z=0) = \boldsymbol{Y}_{\mathrm{coflow}},
\qquad
\boldsymbol{Y}(Z=1) = \boldsymbol{Y}_{\mathrm{fuel}}.
\end{equation}
The corresponding mass-fraction values obtained from the equilibrium evaluation are
\begin{align*}
Y_{\mathrm{H_2}}(Z=0) &= 0.0000, & Y_{\mathrm{H_2}}(Z=1) &= 0.0234, \\
Y_{\mathrm{O_2}}(Z=0) &= 0.1709, & Y_{\mathrm{O_2}}(Z=1) &= 0.0000, \\
Y_{\mathrm{H_2O}}(Z=0) &= 0.0645, & Y_{\mathrm{H_2O}}(Z=1) &= 0.0000, \\
Y_{\mathrm{N_2}}(Z=0) &= 0.7646, & Y_{\mathrm{N_2}}(Z=1) &= 0.9766.
\end{align*}

\subsection{Training Setup}
For all results reported in this paper, a global random seed of $1$ is used for every training run. For the residual loss scaling defined in \eqref{eq:res-weighting}, we use the default hyperparameter value $\lambda = 1$ and do not perform additional tuning. Training is always preformed on a time window $t \in [0, 0.02]$ in which most of the reaction has already taken place.

For the baseline comparisons reported in Table~\ref{table:forward-results}, all methods share the same network and training settings, except for the differences explicitly described below. In the vanilla PINN baseline, we do not employ the residual scaling defined in \eqref{eq:res-weighting}, the mass conservation constraint in \eqref{eq:mass-conservation}, nor the hard constraint on the inert species mass fraction. 

For the parametrized problem, the vanilla baseline does not use the parameterized architecture of \cite{pmlr-v235-cho24b}, but instead employs a fully connected neural network with four hidden layers of 70 units each, resulting in the same depth and a comparable number of trainable parameters. 

For causality-based training, we retain all components of the proposed framework except for the residual scaling based on the reaction term. The remaining modifications are orthogonal to the causality-based weighting strategy and are therefore kept fixed to ensure a fair comparison. The causality-based loss function is defined as
\begin{equation}
\mathcal{L}_{\mathrm{cf}}
=
\sum_{k=1}^{N_t}
w_k \, \mathcal{L}_f(t_k),
\end{equation}
with weights
\begin{equation}
w_k
=
\exp\!\left(
-\gamma
\sum_{m=1}^{k-1}
\mathcal{L}_f(t_m)
\right).
\end{equation}

The hyperparameter $\gamma$ is manually tuned on the parametrized PDE case and is found to have little influence on training performance within the tested range $\gamma \in [10^{-4}, 1]$. Consequently, we fix $\gamma = 1$ for all experiments involving causality-based training.

\paragraph{Inverse Problem}
As outlined in the main text, inverse problem experiments aim to infer the unknown strain rate from observed solution data. These experiments are conducted both in the noise-free setting and in the presence of additive noise. When noise is introduced, synthetic observations are generated by perturbing each species mass fraction independently according to
\begin{equation}
    \tilde{Y}_j
    =
    Y_j
    +
    \epsilon \, \eta_j,
    \qquad
    \eta_j \sim \mathcal{N}\!\left(0,\, \mathrm{std}(Y_j)\right),
\end{equation}
where $Y_j$ denotes the mass fraction of species $j$, $\epsilon$ is the prescribed noise level, and $\mathrm{std}(Y_j)$ is the standard deviation of species $j$ computed over the entire dataset. This construction ensures that the noise magnitude is scaled appropriately for each species based on its variability.

\section{Additional Results}

\begin{figure}[ht]
  \vskip 0.2in
  \begin{center}
    \centerline{\includegraphics[width=0.5\textwidth]{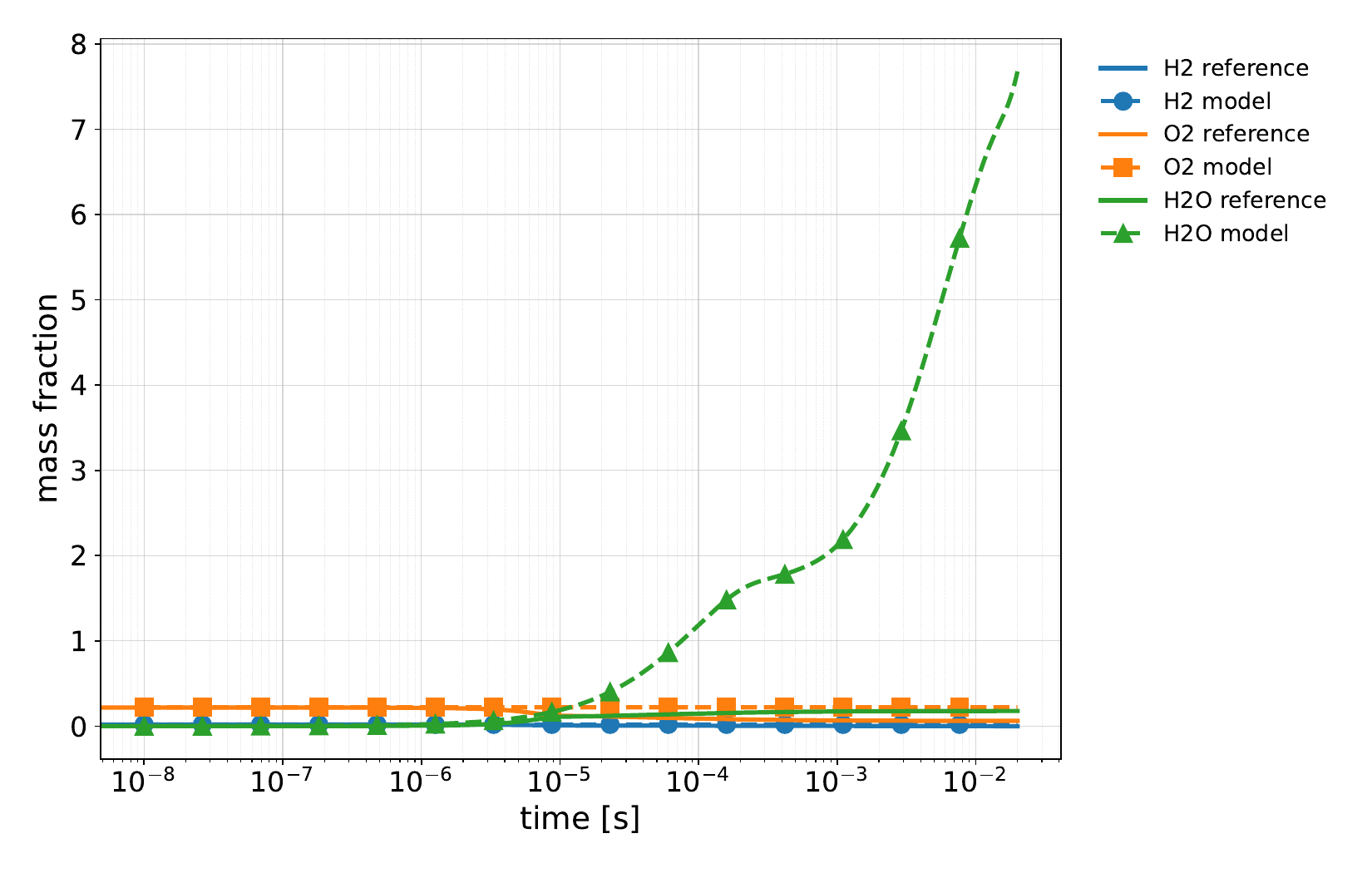}}
    \caption{
      Vanilla PINN solution for the initial value problem ODE. The network fails to capture the rapid transients induced by stiff reaction dynamics. In this plot we included the mass fraction of $N_2$ as it is predicted by the network in the vanilla PINN case.
    }
    \label{fig:IVP-forward-vanilla}
  \end{center}
\end{figure}

\begin{figure}[ht]
  \vskip 0.2in
  \begin{center}
    \centerline{\includegraphics[width=0.5\textwidth]{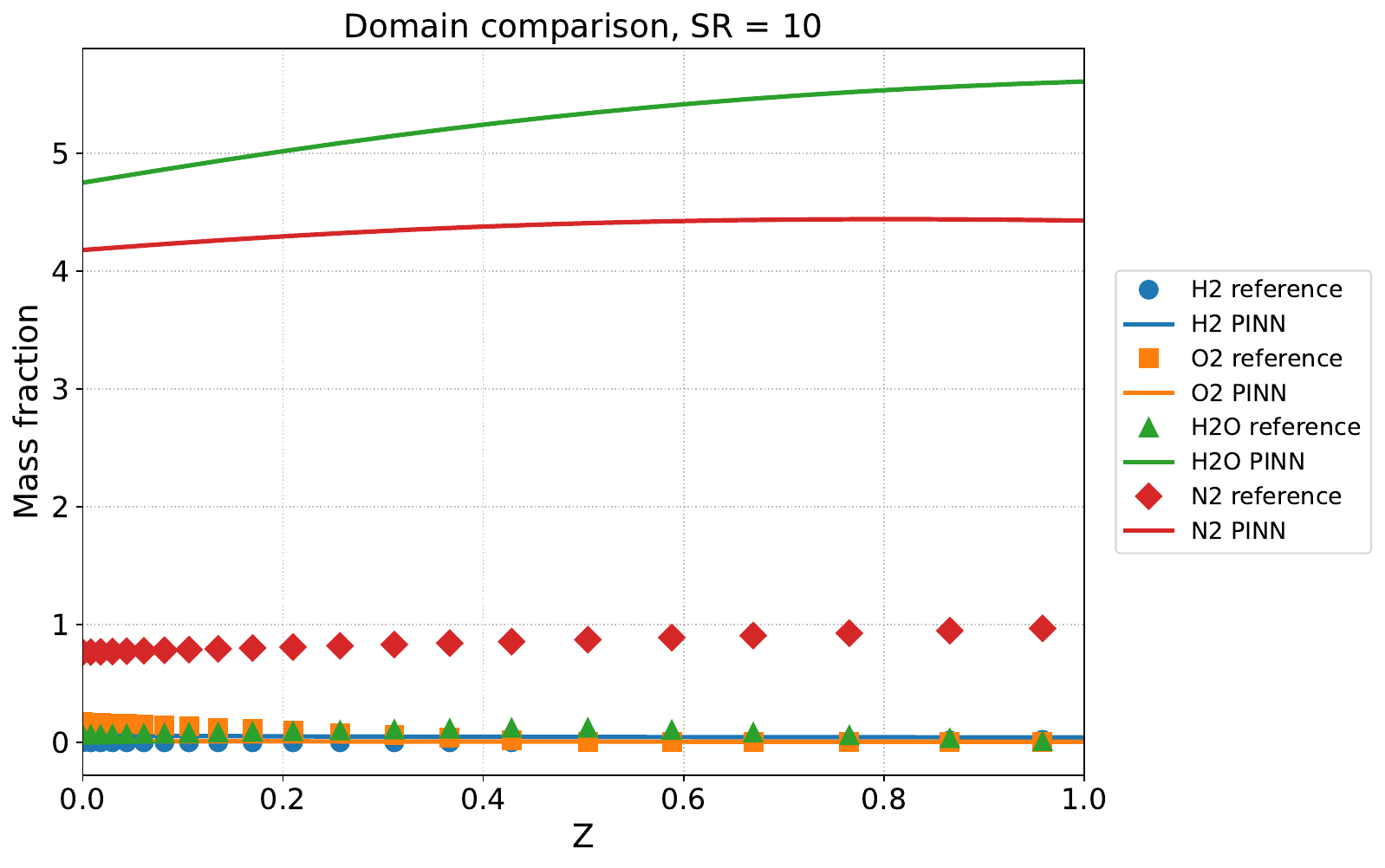}}
    \caption{
      Vanilla PINN solution for the boundary value problem ODE. The predicted species profiles deviate significantly from the physically admissible solution.
    }
    \label{fig:BVP-forward-vanilla}
  \end{center}
\end{figure}

\begin{figure}[ht]
  \vskip 0.2in
  \begin{center}
    \centerline{\includegraphics[width=0.45\textwidth]{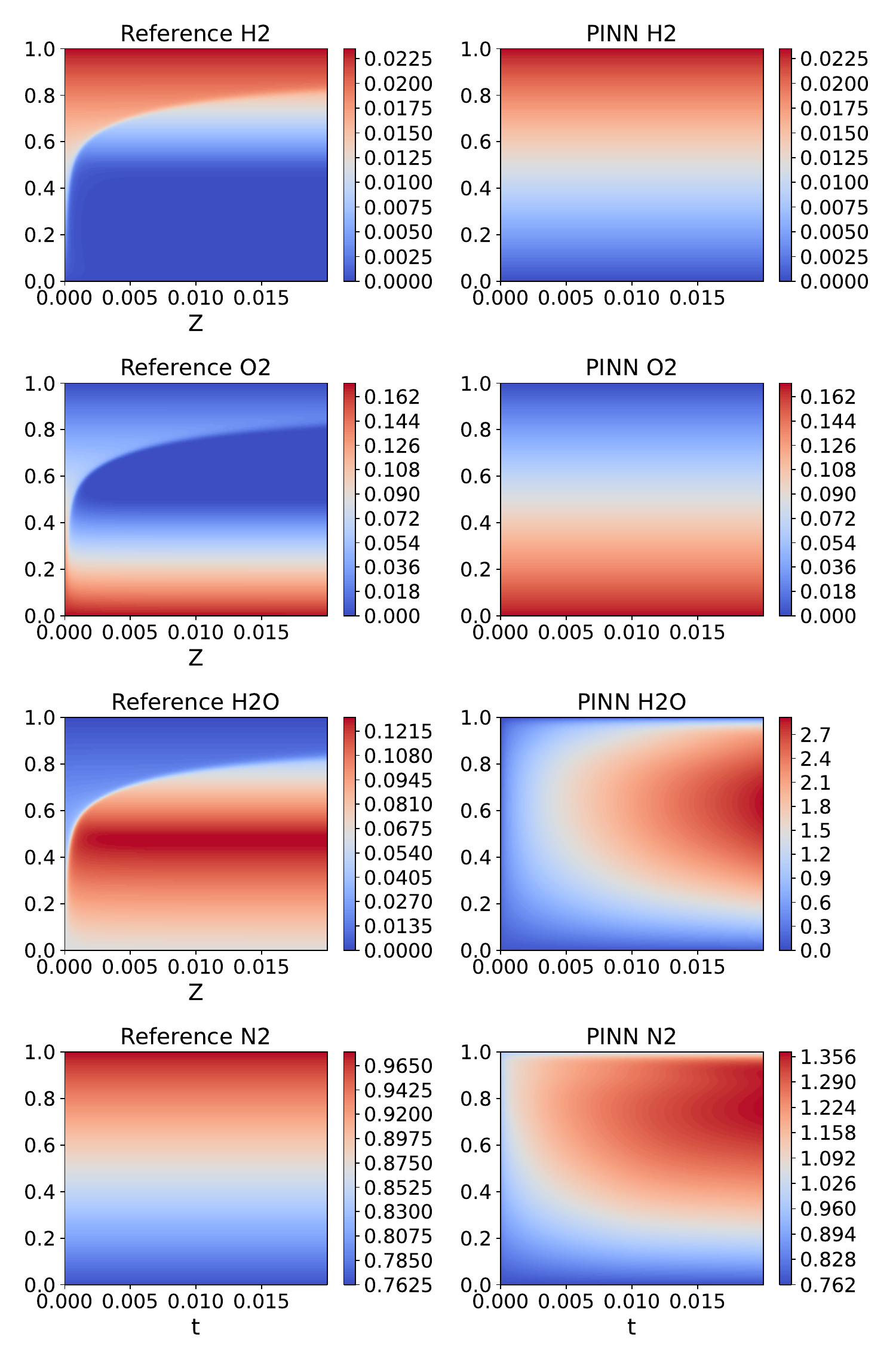}}
    \caption{
      Vanilla PINN solution for the reaction--diffusion PDE at $\alpha = 1$. The model fails to recover the correct spatiotemporal species evolution, demonstrating the inadequacy of standard PINN formulations in this stiff regime.
    }
    \label{fig:PDE-forward-vanilla}
  \end{center}
\end{figure}

Figures~\ref{fig:IVP-forward-vanilla}, \ref{fig:BVP-forward-vanilla}, and \ref{fig:PDE-forward-vanilla} illustrate the behavior of a vanilla physics-informed neural network applied to the initial value problem ODE, the boundary value problem ODE, and the reaction--diffusion flamelet PDE, respectively. In all cases, training without the chemistry-aware modifications introduced in this work fails to converge to a physically meaningful solution and results in severe inaccuracies across the domain. These qualitative failures highlight the necessity of domain-specific stabilization strategies when applying PINNs to extremely stiff reactive systems.

\begin{figure}[ht]
  \centering
  \includegraphics[width=0.48\linewidth]{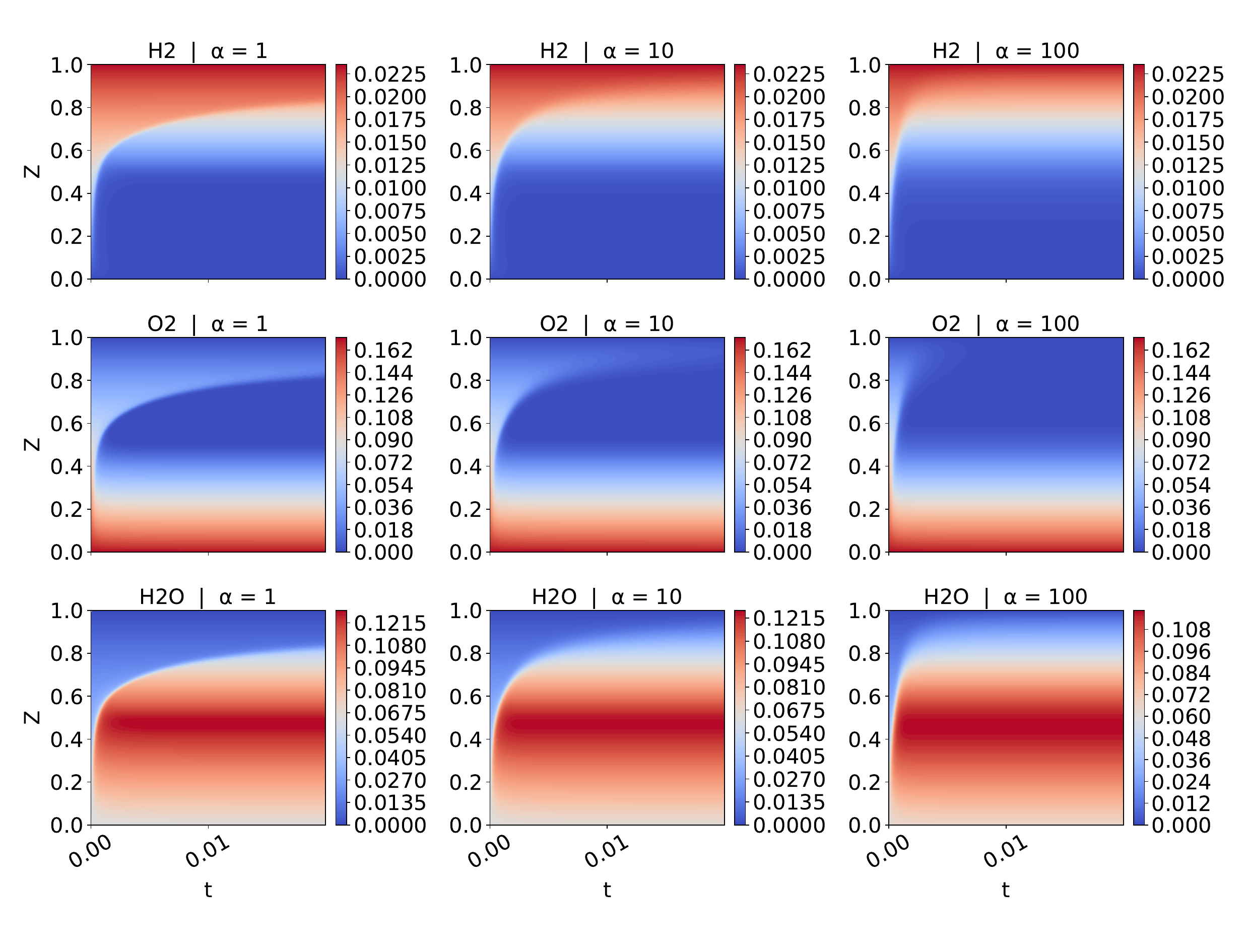}
  \hfill
  \includegraphics[width=0.48\linewidth]{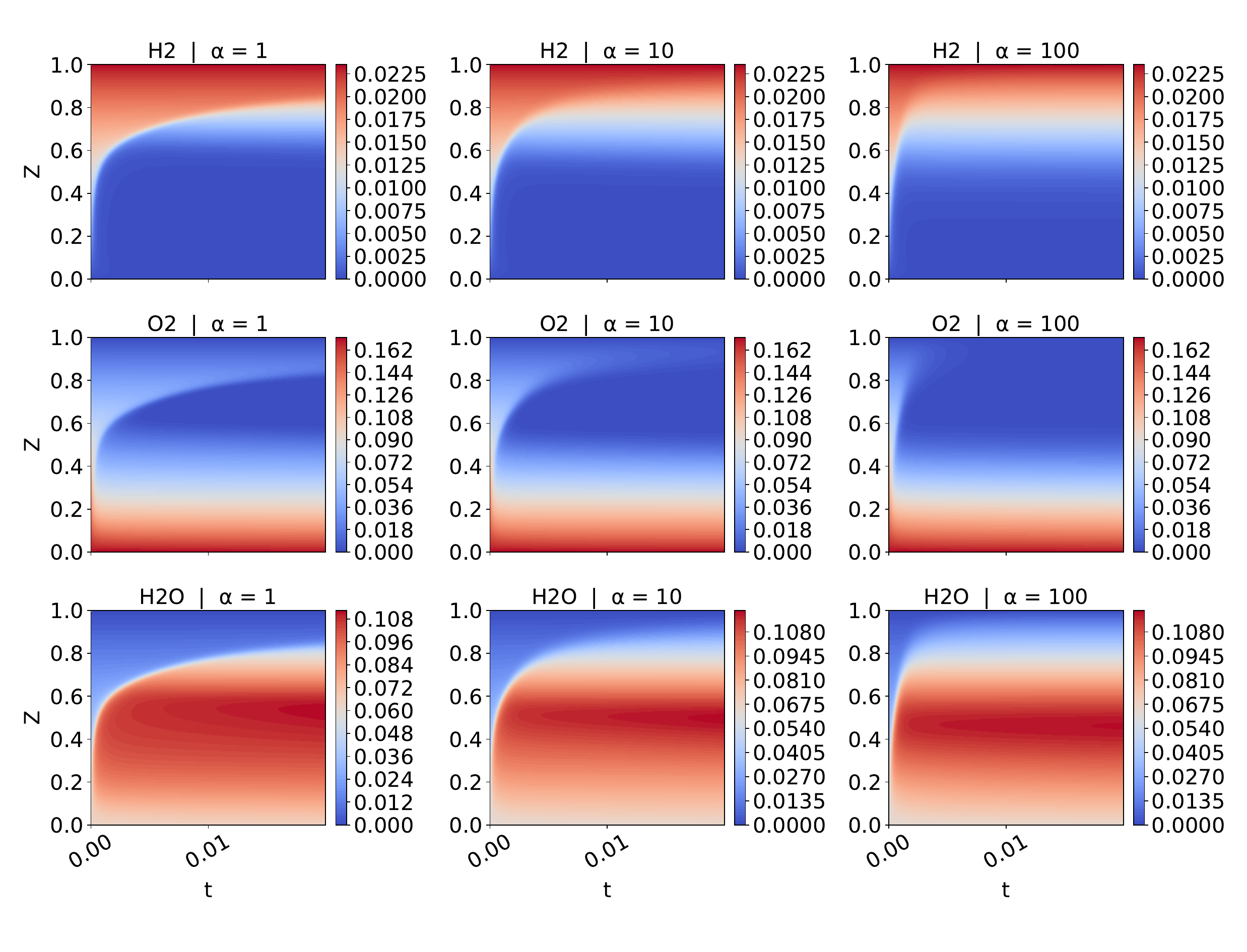}
  \caption{
    Numerical reference solutions (left) and parameterized PINN predictions (right) across multiple strain rates. The proposed framework accurately captures the continuous dependence of the solution on the strain rate.
  }
  \label{fig:parametrized-PDE-comparison}
\end{figure}

Figure~\ref{fig:parametrized-PDE-comparison} compares numerical reference solutions with predictions obtained using the proposed parameterized PINN across multiple strain rates. A single trained model accurately represents a continuous family of flamelet solutions over a wide range of strain rates, demonstrating its ability to capture the parametric structure of the solution manifold.

\end{document}